\pdfoutput=1

\documentclass[11pt]{article}

\usepackage{dingbat}
\usepackage[preprint]{acl}
\usepackage{algorithm}
\usepackage{algorithmic}
\usepackage{dingbat}
\usepackage{times}
\usepackage{latexsym}
\usepackage{amsmath}
\usepackage{pifont}
\usepackage[T1]{fontenc}

\usepackage[utf8]{inputenc}

\usepackage{microtype}

\usepackage{inconsolata}

\usepackage{amsmath}
\usepackage{amssymb}
\usepackage{amsthm}
\usepackage{txfonts}
\usepackage{graphicx}
\usepackage{microtype}
\usepackage{graphicx}
\usepackage{subfigure}
\usepackage{subcaption}
\usepackage{booktabs} 
\usepackage{multicol}
\usepackage{multirow}
\usepackage{makecell}
\usepackage{pgfplots}
\pgfplotsset{compat=newest}
\usepackage{graphicx}
\usepackage{caption}
\usepackage{enumitem}
\usepackage{pifont}
\usepackage{xcolor}
\usepackage{txfonts}
\usepackage{tabularx}
\definecolor{ourRed}{RGB}{209,41,32} %
\definecolor{ourBlue}{RGB}{38,159,255} %
\title{From Misleading Queries to Accurate Answers:\\A Three-Stage Fine-Tuning Method for LLMs}
\author{
Guocong Li$^{1}$, Weize Liu$^{1}$, \
Yihang Wu$^{1}$,  Ping Wang$^{2}$, \\
\bf
Shuaihan Huang$^{1}$, Hongxia Xu$^{1,3}$\thanks{\, Corresponding authors.},\ 
Jian Wu$^{1}$\footnotemark[1]\\
$^{1}$Zhejiang University~~~
$^{2}$Renmin University of China\\
$^{3}$AI Research Center, WeDoctor Cloud\\
\texttt{guocong\_li@163.com} \ \ \ \texttt{weizeliu1115@gmail.com} \\
\texttt{\{einstein,wujian2000\}@zju.edu.cn}
}

\begin{document}
\maketitle
\begin{abstract}
Large language models (LLMs) exhibit excellent performance in natural language processing (NLP), but remain highly sensitive to the quality of input queries, especially when these queries contain misleading or inaccurate information. Existing methods focus on correcting the output, but they often overlook the potential of improving the ability of LLMs to detect and correct misleading content in the input itself. In this paper, we propose a novel three-stage fine-tuning method that enhances the ability of LLMs to detect and correct misleading information in the input, further improving response accuracy and reducing hallucinations. Specifically, the three stages include (1) training LLMs to identify misleading information, (2) training LLMs to correct the misleading information using built-in or external knowledge, and (3) training LLMs to generate accurate answers based on the corrected queries. To evaluate our method, we conducted experiments on three datasets for the hallucination detection task and the question answering~(QA) task, as well as two datasets containing misleading information that we constructed. The experimental results demonstrate that our method significantly improves the accuracy and factuality of LLM responses, while also enhancing the ability to detect hallucinations and reducing the generation of hallucinations in the output, particularly when the query contains misleading information. \footnote{Codes and data are available at: \url{https://github.com/cong03/FMQAA}.}
\end{abstract}

\begin{figure}[ht]
    \centering
    \centerline{\includegraphics[width=0.5\textwidth, trim=0in 0in 0in 0in, clip]{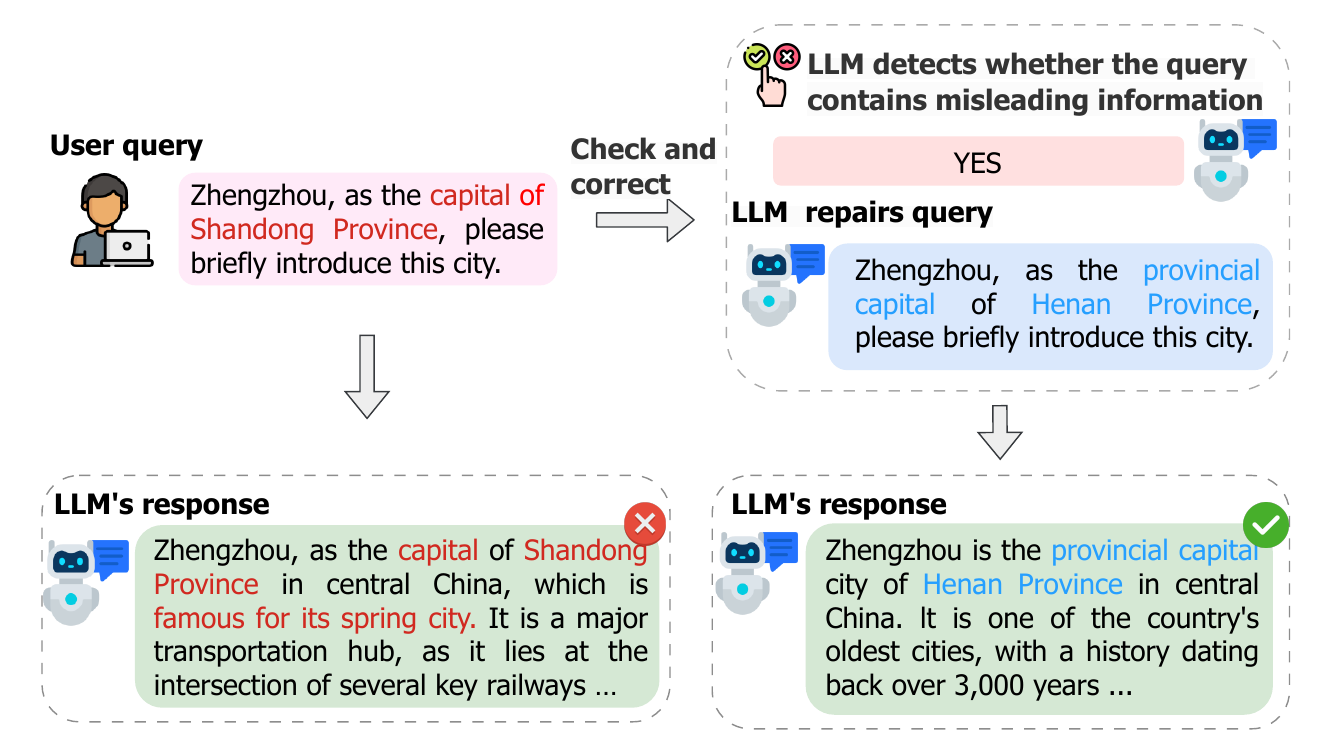}} %
    \small
    \caption{When LLMs directly answer a query containing misleading information~(Left side), they produce incorrect responses. However, when LLMs identify and correct the misleading information in the query before generating a response, they correctly answer it ~(Right side). The misleading information is marked in \textcolor{ourRed}{red}, while the corresponding corrections are marked in \textcolor{ourBlue}{blue}.} %
    \label{fig:image1}
\end{figure}

\section{Introduction}
LLMs have achieved significant success in the field of NLP, particularly excelling in NLG, where they are capable of generating coherent and reasonable responses based on user queries~\citep{li2024leveraging,xuanfan2023systematic}. Although LLMs have made significant progress in generation ability, they are highly sensitive to the quality of the input, and the quality of the generated responses is highly susceptible~\citep{pan2023on,wu2024easily}. Users often provide LLMs with queries that are imprecise or may even contain various forms of misleading information~\citep{nie2020adversarial}. Misleading information refers to inaccurate content within the input that can mislead the model during its reasoning and generation processes. When such information is present in the input, the likelihood of the model generating incorrect responses or hallucinations increases significantly~\citep{yu2024truth,song2024large}. This challenge is particularly prominent in knowledge-intensive domains such as medicine, law, and education, where inputs often contain inaccurate information, and the accuracy of responses is critically important~\citep{barnard2023self}. Therefore, mitigating the impact of misleading information in the input on the performance of LLMs is an urgent and important issue that needs to be addressed.

Current training methods for LLMs primarily focus on correcting the output after generation. An important research direction is the use of the retrieval-augmented generation~(RAG) framework, which leverages external knowledge bases to verify and improve the output~\citep{lewis2020retrieval,gao2023retrieval}. However, RAG typically requires retrieval from large external knowledge bases, a process that can be time-consuming and may not meet the low-latency requirements, especially in real-time application scenarios~\citep{li2024role,jin2024ragcache}. In addition, the self-correction mechanism has also garnered attention, where LLMs iteratively generate and self-evaluate to critique and correct their own outputs~\citep{madaan2024self}. This method has the potential to improve the accuracy and quality of the results, but it also has the drawback that the effectiveness of the correction depends on the model's accuracy in evaluating the results and may require external feedback for assistance~\citep{kamoi2024can}. These existing methods focus on correcting the content generated by LLMs, without considering the correction of misleading information in the input. Although the presence of misleading information in the input may lead to incorrect responses, our proposed method is capable of identifying and detecting errors in the query, subsequently generating the correct answer, as shown in Figure~\ref{fig:image1}. However, no existing work has considered improving the performance of LLMs by enhancing their ability to identify and correct misleading questions.

Therefore, we propose a novel three-stage fine-tuning method that enhances LLMs' ability to detect and correct misleading information in the input. This method enhances the accuracy and robustness of LLMs in processing inputs containing misleading information, while mitigating the negative impact of misleading information on model outputs.

The fine-tuning method that we propose consists of the following three training stages:

\textbf{1) Query detection training: }Training LLMs to determine whether the input query contains misleading information based on built-in knowledge or external knowledge;

\textbf{2) Query correction training: }Training LLMs to combine built-in knowledge or external knowledge to correct the misleading information in the query and generate the correct query;

\textbf{3) Query answering training: }Training LLMs to generate accurate and reliable answers based on the corrected query.

To evaluate the performance of LLMs in answering queries containing misleading information, we construct two misleading datasets, \textit{HaluEval-QA}$_{mis}$ and \textit{CQA}$_{mis}$, based on the \textit{QA} subset of the \textit{HaluEval} dataset and the \textit{CQA} dataset, respectively. Subsequently, we conduct experiments on the misleading datasets, as well as on the three original datasets for the hallucination detection task and the question answering task. The experimental results show that when input contains misleading information, the performance of the trained model using our method is much higher than that of the original model, demonstrating that our method effectively reduces the impact of misleading information in the input on model performance. Furthermore, the model trained using our method also detects that a portion of questions in commonly used standard datasets contain misleading information. Removing these misleading questions can significantly improve the performance of the original model, whereas the model trained with our method consistently maintains stable performance, demonstrating its robustness to misleading information in the input. Regardless of whether the input contains misleading information, our method significantly improves the accuracy and factuality of LLM responses, as well as the ability to detect hallucinations in the responses, while effectively mitigating the hallucinations of the model.

\section{A Fine-Tuning Method to Mitigate the Impact of Misleading Information}

\begin{figure*}[ht]
    \centering
    \includegraphics[width=1\textwidth]{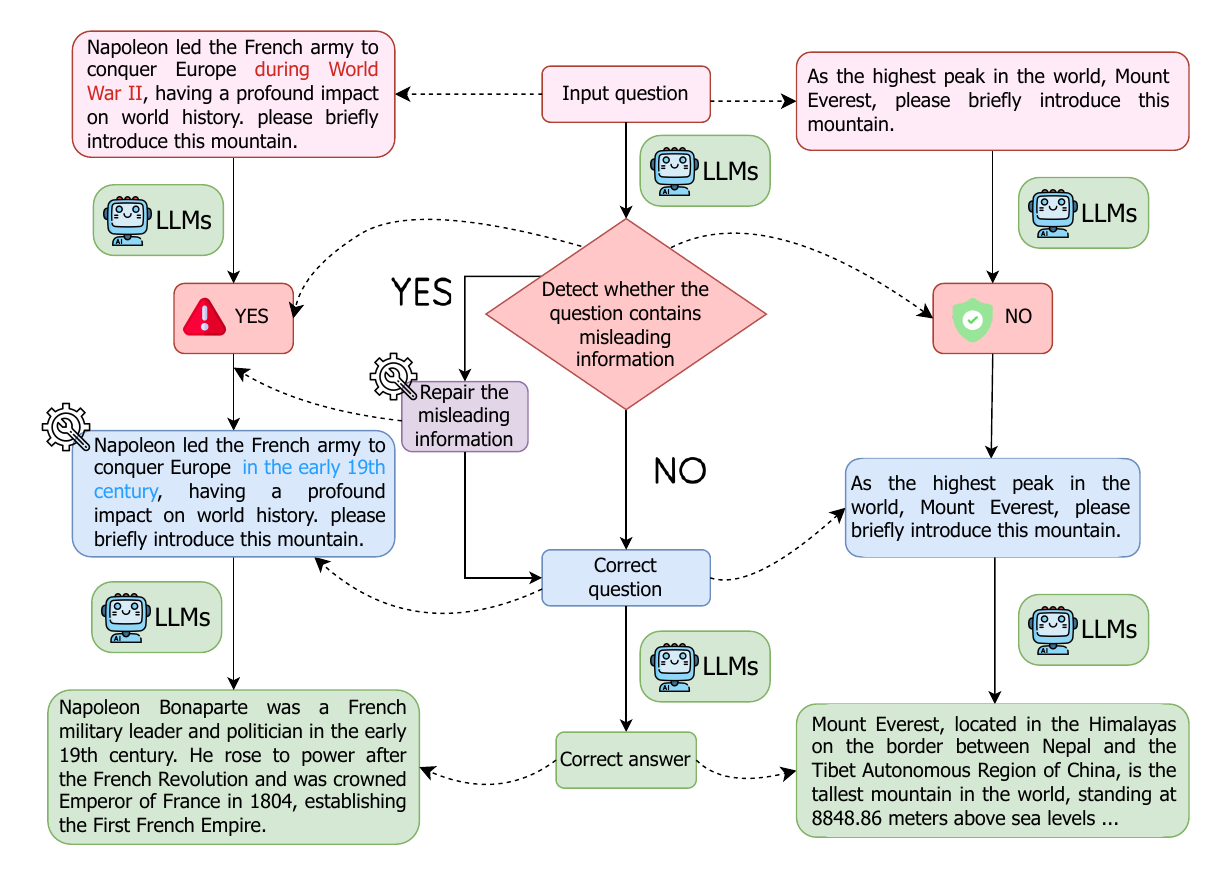}
    \caption{The framework diagram of our method. In the first stage, we train LLMs to determine whether the query contains misleading information. In the second stage, for queries containing misleading information, we train LLMs to correct the misleading information in the query to obtain a correct query. In the third stage, we train LLMs to generate the correct answer based on the corrected query or the original query without misleading information. Both the Correct Query and Correct Answer can be directly derived from the content of the original dataset.}
    \label{fig:image2}
\end{figure*}

The framework of the proposed method is illustrated in Figure~\ref{fig:image2}. It involves training LLMs to sequentially perform the following tasks: detecting misleading information in the input query, correcting the misleading query, and generating accurate answers based on the corrected query, thereby improving the model's accuracy and robustness when handling inputs containing misleading information.

\subsection{Construction of Misleading Query Datasets}

We used the Qwen2.5-72B-Instruct model to generate a dataset containing misleading information based on the original queries. Specifically, we constructed the misleading query dataset \textit{HaluEval-QA}$_{mis}$ by augmenting the queries from the \textit{QA} subset of the \textit{HaluEval} dataset with misleading content, leveraging the knowledge from the original queries in the \textit{HaluEval} dataset. Similarly, based on the \textit{CQA} dataset, we used the Qwen2.5-72B-Instruct model to modify each question in the original dataset to contain misleading information, thereby constructing the misleading query dataset \textit{CQA}$_{mis}$. For each dataset, we designed corresponding prompts to ensure that the newly generated misleading query datasets introduced misleading information while preserving the naturalness and complexity of the original queries. The prompt templates and data examples used are provided in Appendix~\ref{sec:appendixb}. At the same time, to ensure that the generated misleading query datasets are of higher quality, we adopted a batch generation strategy, generating three misleading queries per batch, and filtered the generated data based on the following two criteria:

\paragraph{Sentence Similarity} The similarity between the generated misleading query and the original query is measured using edit distance, which is defined as the minimum number of operations (including insertion, deletion, and substitution) required to transform one string into another~\citep{niu2018word}. Let the original query be $q_{\text{ori}}$ and the generated misleading query be $q_{\text{mis}}$. The sentence similarity $S_{\text{sim}}$ is defined as:
\begin{equation}
S_{\text{sim}} = 1 - \frac{\text{EditDistance}(q_{\text{ori}}, q_{\text{mis}})}{\max(\text{Length}(q_{\text{ori}}), \text{Length}(q_{\text{mis}}))},
\end{equation}
\noindent where $\text{EditDistance}(q_{\text{ori}}, q_{\text{mis}})$ represents the edit distance between $q_{\text{ori}}$ and $q_{\text{mis}}$, and Length denotes the length of the sentence. We retain misleading queries with surface similarity $S_{\text{sim}}$ higher than the threshold $r_{\text{sim}}$ to ensure that the generated queries are close to the original queries.

\paragraph{Answer Error Rate} The answer error rate measures whether the model generates incorrect answers when responding to generated, misleading queries. We input the query into the model $N$ times, record the number of incorrect answers $N_{\text{error}}$, and define the error rate as:
\begin{equation}
E_{\text{error}} = \frac{N_{\text{error}}}{N},
\end{equation}
where $E_{\text{error}}$ denotes the error rate. We only retain the misleading queries with an answer error rate $E_{\text{error}}$ above the threshold $r_{\text{error}}$ to ensure that these queries have a high level of misleading content.

\paragraph{Data Filtering Strategy}
Specifically, for each query, we randomly generate three misleading queries that satisfy $S_{\text{sim}}$ > 0.8 and $E_{\text{error}}$ > 0.5, and then retain the one with the highest value of $S_{\text{sim}} + E_{\text{error}}$. An example of randomly generated misleading queries is presented in Appendix ~\ref{sec:appendixc}. This ensures that the generated misleading queries are both highly similar to the original queries and contain sufficiently misleading information, thereby enabling more effective training and evaluation of LLMs' performance when faced with highly deceptive and subtle misleading queries.

\subsection{Query Detection Training}

Since misleading information in the query can significantly affect the quality of LLMs' responses, the goal of this stage is to train the model to identify whether misleading information exists in the given query.

\paragraph{Input} A query. (If the dataset contains background knowledge, the background knowledge is also input to the model, such as \textit{HaluEval} dataset.)

\paragraph{Output Objective} YES (indicating that the query contains misleading information) or NO (indicating that the query contains no misleading information).

\paragraph{Training Objective} The training objective is to minimize the cross-entropy loss function:

\begin{equation}
\begin{split}
\mathcal{L}_{\text{check}} =& -\frac{1}{N} \sum_{i=1}^{N} \left( y_{\text{true}, i} \log(y_{\text{pred}, i}) \right. \\
&\quad + \left. (1 - y_{\text{true}, i}) \log(1 - y_{\text{pred}, i}) \right)
\end{split},
\end{equation}
where $N$ is the number of samples in each batch, $y_{\text{true}, i}$ is the true label of sample \(i\), with a value of 1 (indicating ``YES'') or 0 (indicating ``NO''), and $y_{\text{pred}, i}$ is the model's predicted probability that the \(i\)-th sample is labeled as ``YES''.

\subsection{Query Correction Training} Queries labeled as containing misleading information (``YES'') in the query detection stage will proceed to the query correction stage. The goal of this stage is to correct the queries by leveraging the knowledge provided by the dataset or the model's internal knowledge, transforming them into versions that do not contain misleading information, thereby improving the query quality.

\paragraph{Input} A misleading query. (If the dataset contains background knowledge, the background knowledge is also input to the model, such as \textit{HaluEval dataset}.)

\paragraph{Output Objective} Corrected Query.

\paragraph{Training Objective} The loss function in the query correction stage uses semantic similarity rather than edit distance, as misleading information primarily affects semantic similarity rather than edit distance~\citep{wu2022semantic}. This approach is used to train the model to generate high-quality corrected queries that are semantically consistent with the original correct queries. In this study, we use Sentence-BERT to obtain semantic vector representations~\citep{reimers2019sentence}. Furthermore, we ensure that corrected queries are semantically and logically consistent with correct queries by minimizing the distance between corrected and correct queries in the semantic space. Mathematically, the loss function is formulated as:

\begin{equation}
\mathcal{L}_{\text{correction}} = - \frac{1}{N} \sum_{i=1}^{N}\left( 1 - \frac{\mathbf{u}_i \cdot \mathbf{v}_i}{\|\mathbf{u}_i\| \|\mathbf{v}_i\|} \right),
\end{equation}
where $N$ is the number of training samples in each batch, and $\mathbf{u}_i$ and $\mathbf{v}_i$ are the semantic vector representations of the \(i\)-th correct query and corrected query, respectively.

\subsection{Query Answering Training} The goal of this stage is to generate accurate and reliable answers based on the corrected query or the original query.

\paragraph{Input} The corrected query or the original query.

\paragraph{Output Objective} Correct Answers.

\paragraph{Training Objective} The model-generated answers should be aligned with the ground truth answers. The training objective is to minimize the cross-entropy loss function. Given the input sequence of the \(i\)-th data sample as $X^{(i)} = x_1^{(i)}, x_2^{(i)}, \dots, x_{T_i}^{(i)}$, the loss function is defined as:
\begin{equation}
\begin{split}
& \mathcal{L}_{\text{generation}} =  \\
& -\frac{1}{N} \sum_{i=1}^{N} \left( \frac{1}{T_i} \sum_{t=1}^{T_i} \log P(y_t^{(i)} | x_1^{(i)}, x_2^{(i)}, \dots, x_{t-1}^{(i)}) \right)
\end{split},
\end{equation}

\noindent where $N$ is the number of training samples in each batch,  $P(y_t^{(i)} | x_1^{(i)}, x_2^{(i)}, \dots, x_{t-1}^{(i)})$ denotes the probability that the model predicts the ground truth label $y_t^{(i)}$ at position \(t\) in the \(i\)-th data sample and $T_i$ represents the sequence length of the \(i\)-th data sample.

\subsection{Unified Framework and Training }To ensure consistency across the query detection, correction, and answer generation processes, we carefully train the model for the next stage based on the model trained in the previous stage, ultimately forming a seamless pipeline. Through this three-stage training strategy, the model effectively learns to detect and correct misleading information in queries and generate accurate responses. This approach significantly mitigates the impact and potential harm of misleading information on the model's performance.

\section{Experiments}

\begin{table*}[ht]
\small
\setlength\tabcolsep{4pt}
\centering
\begin{tabular}{llcccc}
\toprule
\multirow{2}{*}{\textbf{Instruction Model}} & \multirow{2}{*}{\textbf{Method}} & \multicolumn{4}{c}{\textbf{\textit{HaluEval}}}  \\
\cmidrule(lr){3-6} 
 & & \textbf{\textit{Sum}} & \textbf{\textit{Dia}} & \textbf{\textit{QA}} & \textbf{\textit{HaluEval-QA}$_{mis}$}
 
 \\
\midrule

\multirow{3}{*}{Qwen2-7B}
& Original  & $51.5 \pm 0.8$ & $70.3 \pm 0.4$ & $54.5 \pm 1.3$ & $51.1 \pm 0.6$ \\
& SFT & $54.2 \pm 1.8$ & $71.9 \pm 0.5$ & $62.3 \pm 1.1$ & $58.6 \pm 1.3$ \\
& Ours & \textbf{68.1 $\pm$ 1.7} & \textbf{72.1 $\pm$ 1.3} & \textbf{68.9 $\pm$ 1.6} & \textbf{68.0 $\pm$ 1.0} \\
\midrule
\multirow{3}{*}{Qwen2.5-14B}
& Original  & $73.2 \pm 0.6$ & $70.5 \pm 0.9$ & $62.7 \pm 1.7$ & $54.3 \pm 0.8$  \\
& SFT & $73.8 \pm 0.5$ & $71.3 \pm 0.6$ & $68.2 \pm 1.3$ & $60.7 \pm 1.0$  \\
& Ours & \textbf{76.3 $\pm$ 0.8} & \textbf{72.2 $\pm$ 0.4} & \textbf{73.2 $\pm$ 1.5} & \textbf{71.3 $\pm$ 1.6}  \\
\midrule
\multirow{3}{*}{Llama3.2-3B}
& Original  & $27.2 \pm 2.3$ & $49.9 \pm 0.5$ & $49.5 \pm 1.2$ & $46.2 \pm 0.7$ \\
& SFT & $35.0 \pm 1.6$ & $53.2 \pm 1.3$ & $51.9 \pm 1.0$ & $48.1 \pm 0.4$ \\
& Ours & \textbf{57.4 $\pm$ 2.3} & \textbf{60.8 $\pm$ 2.1} & \textbf{53.9 $\pm$ 0.8} & \textbf{52.5 $\pm$ 0.9} \\
\midrule
\multirow{3}{*}{Llama3-8B}
& Original  & $50.0 \pm 1.8$ & $57.3 \pm 1.4$ & $51.8 \pm 0.9$ & $49.6 \pm 0.6$ \\
& SFT & $53.8 \pm 0.8$ & $59.1 \pm 1.3$ & $57.3 \pm 1.3$ & $53.2 \pm 1.0$ \\
& Ours & \textbf{62.7 $\pm$ 1.5} & \textbf{68.6 $\pm$ 0.6} & \textbf{67.7 $\pm$ 1.8} & \textbf{66.9 $\pm$ 1.4} \\
\bottomrule
\end{tabular}
\caption{Performance (mean $\pm$ standard deviation, \%) comparison on \textit{HaluEval} dataset. In this case, Original refers to the model without any additional processing, while supervised fine-tuning~(SFT) refers to the model that undergoes direct fine-tuning.}
\label{tab:table1}
\end{table*}

\subsection{Setup} We conducted comprehensive experiments on the hallucination detection task (\textit{HaluEval}), question answering tasks (\textit{CQA} and \textit{TruthfulQA}), and the misleading query datasets we constructed (\textit{HaluEval-QA}$_{mis}$ and \textit{CQA}$_{mis}$). In all experiments, we report the mean results and standard deviations over three random runs. The experimental configuration and dataset usage details are in Appendix~\ref{sec:appendixa}.

\paragraph{Original Datasets} We conduct the construction of the misleading datasets and the evaluation of the model on three datasets:
\begin{itemize}
    \item The \textit{HaluEval} dataset~\citep{li2023halueval} includes the \textit{QA}, \textit{Summarization}, and \textit{Dialogue} subsets, which are used to test the model's ability to identify hallucinations and misleading information.
    \item The \textit{CQA} dataset~\citep{talmor2019commonsenseqa} is used to evaluate the model's accuracy in answering common sense questions.
    \item The \textit{TruthfulQA} dataset~\citep{lin2022truthfulqa} is used to assess the truthfulness and hallucination in the generated answers.
\end{itemize}

\paragraph{Construction of Training Datasets} Our training dataset consists of 4,000 samples from the \textit{QA} subset of \textit{HaluEval} and the corresponding 4,000 samples from the \textit{HaluEval-QA}$_{mis}$ dataset, which contains misleading information. This dataset is designed to provide diverse training samples that include both genuine and misleading queries. The specific construction process is in Appendix~\ref{sec:appendixb}.

\begin{table*}[ht]
\small
\centering
\begin{tabular}{llccccc}
\toprule
\multirow{2}{*}{\textbf{Instruction Model}} & \multirow{2}{*}{\textbf{Method}} & \textbf{\textit{CQA}} & \textbf{\textit{CQA}$_{mis}$} & \multicolumn{3}{c}{\textbf{\textit{TruthfulQA}}} \\
\cmidrule(lr){3-3}
\cmidrule(lr){4-4}
\cmidrule(lr){5-7}
 & & \textbf{ACC} & \textbf{ACC} & \textbf{BLEURT} & \textbf{BLEU} & \textbf{ROUGE} \\
\midrule
\multirow{3}{*}{Qwen2-7B}
& Original & 70.4 $\pm$ 0.8 & 68.8 $\pm$ 0.8 & 0.636 $\pm$ 0.013 & 0.543 $\pm$ 0.006 & 0.570 $\pm$ 0.004 \\
& SFT & 72.1 $\pm$ 0.7 & 69.9 $\pm$ 1.1 & 0.633 $\pm$ 0.011 & \textbf{0.562 $\pm$ 0.002} & 0.577 $\pm$ 0.007 \\
& Ours & \textbf{78.8 $\pm$ 1.6} & \textbf{74.9 $\pm$ 0.5} & \textbf{0.678 $\pm$ 0.006} & 0.551 $\pm$ 0.003 & \textbf{0.589 $\pm$ 0.006} \\
\midrule
\multirow{3}{*}{Qwen2.5-14B}
& Original  & 80.9 $\pm$ 0.8 & 71.6 $\pm$ 0.9 & 0.709 $\pm$ 0.019 & 0.582 $\pm$ 0.004 & 0.609 $\pm$ 0.009 \\
& SFT & 81.4 $\pm$ 0.5 & 73.3 $\pm$ 1.6 & 0.721 $\pm$ 0.005 & 0.577 $\pm$ 0.008 & 0.613 $\pm$ 0.007 \\
& Ours & \textbf{83.1 $\pm$ 0.7} & \textbf{78.6 $\pm$ 1.3} & \textbf{0.748 $\pm$ 0.008} & \textbf{0.615 $\pm$ 0.011} & \textbf{0.642 $\pm$ 0.013} \\
\midrule
\multirow{3}{*}{Llama3.2-3B}
& Original  & 65.1 $\pm$ 0.9 & 62.1 $\pm$ 0.6 & 0.613 $\pm$ 0.006 & 0.517 $\pm$ 0.005 & 0.527 $\pm$ 0.005 \\
& SFT & 66.2 $\pm$ 1.0 & 63.7 $\pm$ 0.7 & 0.617 $\pm$ 0.005 & 0.523 $\pm$ 0.004 & 0.521 $\pm$ 0.008 \\
& Ours & \textbf{68.5 $\pm$ 0.4} & \textbf{67.5 $\pm$ 0.5} & \textbf{0.632 $\pm$ 0.008} & \textbf{0.534 $\pm$ 0.007} & \textbf{0.556 $\pm$ 0.007} \\
\midrule
\multirow{3}{*}{Llama3-8B}
& Original  & 70.8 $\pm$ 1.1 & 64.9 $\pm$ 0.5 & 0.625 $\pm$ 0.011 & 0.489 $\pm$ 0.012 & 0.553 $\pm$ 0.004 \\
& SFT & 69.3 $\pm$ 0.5 & 64.3 $\pm$ 0.4 & 0.607 $\pm$ 0.013 & 0.503 $\pm$ 0.005 & 0.562 $\pm$ 0.008 \\
& Ours & \textbf{76.4 $\pm$ 1.3} & \textbf{70.0 $\pm$ 0.8} & \textbf{0.668 $\pm$ 0.009} & \textbf{0.528 $\pm$ 0.007} & \textbf{0.586 $\pm$ 0.006} \\
\bottomrule
\end{tabular}
\caption{Performance (mean $\pm$ standard deviation, \%) comparison on \textit{CQA}, \textbf{\textit{CQA}$_{mis}$} and \textit{TruthfulQA} dataset.}
\label{tab:table2}
\end{table*}

\subsection{Main Results}
We conducted experiments on models from the Qwen~\citep{yang2024qwen2} and Llama~\citep{dubey2024llama} families. The experimental results for the hallucination detection task are listed in Table~\ref{tab:table1}, and the experimental results for the question answering task are listed in Table~\ref{tab:table2}.

\paragraph{The Impact of Misleading Information in the Input on the Response} For all the original models (those not trained with our framework), the accuracy on the \textit{HaluEval-QA}$_{mis}$ and \textit{CQA}$_{mis}$ datasets is significantly lower than that on the \textit{QA} and \textit{CQA} datasets. This demonstrates that the presence of misleading information in the input can significantly affect the accuracy and performance of the model's responses. After SFT, the model shows some improvement in performance on both the original dataset and the misleading query datasets. However, the accuracy on the misleading query datasets \textit{HaluEval-QA}$_{mis}$ and \textit{CQA}$_{mis}$ remains significantly lower than that on the original \textit{QA} and \textit{CQA} datasets. However, after training with our method, the model's performance on the misleading query datasets shows a significant reduction in the gap compared to the \textit{QA} and \textit{CQA} datasets. This indicates that our proposed method effectively mitigates the performance degradation caused by misleading information in the input.

\paragraph{Performance on the Hallucination Detection Task} We tested the model's performance on the hallucination detection task using the \textit{QA}, \textit{Summarization}, and \textit{Dialogue} subsets of the \textit{HaluEval} dataset, as well as the \textit{HaluEval-QA}$_{mis}$ dataset. The experimental results show that the model trained by our method significantly outperforms the original model as well as the model after SFT on all datasets. This demonstrates that our method enhances the model's ability to detect misleading information.

\paragraph{Performance on the Question Answering Task} We evaluated the model on the commonsense question answering task using the \textit{CQA} dataset and the \textit{CQA}$_{mis}$ dataset. We found that the model trained with our method showed a significant improvement in accuracy on both datasets, demonstrating that our method can enhance the model's accuracy in question answering tasks. We tested the model's ability to generate truthful responses and avoid producing false information using the \textit{TruthfulQA} dataset. The experimental results show that the model trained with our method achieves improvements in BLEURT\citep{sellam2020bleurt}, BLEU~\citep{papineni2002bleu}, and ROUGE~\citep{lin2004rouge} metrics, demonstrating that our method can effectively reduce hallucinations in responses and improve the factual accuracy of the model's answers.

\paragraph{Scalability Analysis} We selected the 7B and 14B models from the Qwen series, as well as the 3B and 8B models from the Llama series, for experimentation. The experimental results show that, regardless of the model family or size, the performance of the models trained with our method is significantly improved. This indicates that our method is applicable to models of different scales and demonstrates good scalability. Furthermore, on the \textit{QA}, \textit{HaluEval-QA}$_{mis}$, and \textit{Dialogue} datasets, the Qwen-7B model trained with our framework even outperforms the Qwen-14B model that was not trained with our method, further validating the effectiveness and scalability of our approach.

\paragraph{Cross-task Generalisability Analysis} During the model training process, our training data only consisted of 4,000 samples from the \textit{QA} dataset and 4,000 samples from the \textit{HaluEval-QA}$_{mis}$ dataset. However, the model still shows significant performance improvements on other datasets. In comparison, we performed SFT on the model using the same training data. However, the model fine-tuned with standard methods showed significant performance improvement only on the \textit{QA} and \textit{HaluEval-QA}$_{mis}$ datasets, with little to no improvement on other datasets and even a decline in performance. This demonstrates that our method has strong cross-task generalization capabilities.

\subsection{Instruction Following Capability Study}

We evaluated the model's instruction-following capability on the three subtasks of the hallucination detection dataset \textit{HaluEval} and the \textit{HaluEval-QA}$_{mis}$ dataset. In these evaluations, if the model's response does not include either ``YES'' or ``NO'', or contains both, it is marked as failed. Table~\ref{tab:table3} shows the proportion of failed responses for each model in this evaluation task. The models trained with our method exhibit a significant reduction in the failed proportion across all datasets, indicating that our method significantly improves the model's instruction-following capability.

\begin{table}[ht]
\setlength\tabcolsep{2.5pt}
\small
\centering
\begin{tabular*}{0.47\textwidth}{llcccc}
\toprule
\multirow{2}{*}{\textbf{\parbox{1.3cm}{Instruction \\ Model}}} & \multirow{2}{*}{\textbf{Method}} & \multicolumn{4}{c}{\textbf{\textit{HaluEval}}}  \\
\cmidrule(lr){3-6} 

  & & \textbf{\textit{Sum}} & \textbf{\textit{Dia}} & \textbf{\textit{QA}} & \textbf{\textit{HaluEval-QA}$_{mis}$} 
\\
\midrule
\multirow{3}{*}{Llama3.2-3B}
& Original & 13.67 & 1.96 & 3.54 & 5.67  \\
& SFT  & 8.73 & 2.17 & 2.36 & 2.86 \\
& Ours & \textbf{3.55}  & \textbf{1.24} & \textbf{1.09} & \textbf{2.09} \\
\midrule
\multirow{3}{*}{Llama3-8B}
& Original  & 2.61  & 0.63 & 1.03 & 1.75 \\
& SFT  & 2.17  & 0.58 & 0.95 & 1.42 \\
& Ours  & \textbf{0.36}  & \textbf{0.43} & \textbf{0.69} & \textbf{0.78} \\
\bottomrule
\end{tabular*}
\caption{Performance comparison of failure rate (\%) on HaluEval dataset. A smaller failure rate (\%) indicates a stronger ability of LLMs to follow instructions.}
\label{tab:table3}
\end{table}

\begin{table*}[ht]
\small
\setlength\tabcolsep{4pt}
\centering
\begin{tabular}{llcccccc}
\toprule
\multirow{2}{*}{\textbf{Instruction Model}} & \multirow{2}{*}{\textbf{Method}} & \multicolumn{4}{c}{\textbf{\textit{HaluEval}}}  & \multicolumn{2}{c}{\textbf{\textit{CQA}}}\\
\cmidrule(lr){3-6} \cmidrule(lr){7-8}
 & & \textbf{\textit{Sum}} & \textbf{\textit{Dia}} & \textbf{\textit{QA}} & \textbf{\textit{HaluEval-QA}$_{mis}$}  & \textbf{\textit{CQA}}  & \textbf{\textit{CQA}$_{mis}$} \\
\midrule

\multirow{4}{*}{Llama-3.2-3B}
& Original  & $27.2 \pm 2.3$ & $49.9 \pm 0.5$  & $49.5 \pm 1.2$ & $46.2 \pm 0.7$ & $65.1 \pm 0.9$ & $62.1 \pm 0.6$ \\
& Stage 1 &$52.8 \pm 1.0$ & $56.4 \pm 1.5$ &  $50.1 \pm 1.0$ & $48.4 \pm 0.6$ & $67.6 \pm 0.6$ & $66.1 \pm 0.9$ \\
& Stage 2 & $53.2 \pm 0.7$ & $57.2 \pm 1.3$  & $50.5 \pm 0.7$ & $48.7 \pm 0.4$ & $68.1 \pm 0.8$ & $66.5 \pm 0.4$ \\
& Stage 3 & \textbf{57.4 $\pm$ 2.3} & \textbf{60.8 $\pm$ 2.1} & \textbf{53.9 $\pm$ 0.8} & \textbf{52.5 $\pm$ 0.9} & \textbf{68.5 $\pm$ 0.4} & \textbf{67.5 $\pm$ 0.5} \\
\midrule

\multirow{4}{*}{Llama3-8B}
& Original  & $50.0 \pm 1.8$ & $57.3 \pm 1.4$ & $51.8 \pm 0.9$ & $49.6 \pm 0.6$ & $70.8 \pm 1.1$ & $64.9 \pm 0.5$ \\
& Stage 1 & $60.1 \pm 0.6$ & $65.3 \pm 1.2$  & $63.1 \pm 1.2$ & $62.1 \pm 1.2$ & $73.5 \pm 1.2$ & $67.4 \pm 1.1$ \\
& Stage 2 & $60.5 \pm 0.8$ & $65.7 \pm 0.8$ & $63.6 \pm 0.9$ & $63.0 \pm 0.7$ & $74.8 \pm 1.9$ & $68.9 \pm 0.7$ \\
& Stage 3 & \textbf{62.7 $\pm$ 1.5} & \textbf{68.6 $\pm$ 0.6} & \textbf{67.7 $\pm$ 1.8} & \textbf{66.9 $\pm$ 1.4} & \textbf{76.4 $\pm$ 1.3} & \textbf{70.0 $\pm$ 0.8} \\
\bottomrule
\end{tabular}
\caption{Ablation study on \textbf{\textit{HaluEval} and \textit{CQA} dataset.} Original refers to the original model, Stage 1 refers to the model after query detection training, Stage 2 refers to the model after further query correction training, and Stage 3 refers to the model after further query answering training again.}
\label{tab:table4}
\end{table*}

\subsection{Ablation Study}
We used the 3B and 8B versions of the Llama models to test the changes in model performance after each stage of training. The results are shown in Table~\ref{tab:table4}.

\paragraph{The Role of the Query Detection Stage} As can be seen from the results in Table~\ref{tab:table4}, the models after query detection training show significant performance improvement on all datasets. Moreover, the larger the model size, the more significant the performance improvement. This verifies that the query detection stage effectively enhances the model's ability to match query details with existing knowledge, thereby enabling it to identify potential misleading information more accurately. Therefore, after fine-tuning with the query detection stage, the model's ability to identify misleading information is significantly enhanced, providing a solid foundation for subsequent hallucination correction and answer generation.

\paragraph{The Role of the Query Correction Stage} 
As can be seen from the results in Table~\ref{tab:table4}, the models, after further query correction training, continue to exhibit performance improvements across all datasets. In Appendix~\ref{sec:appendixd}, we present a case study that illustrates that after this step of training, the model has learned to correct queries containing misleading information.

\paragraph{The Role of the Answer Generation Stage} As can be seen from the results in Table~\ref{tab:table4}, the models after further query answering training show significant improvement both on the hallucination detection task and the question answering task. The main role of the answer generation training is to help the model adapt to the corrected queries, enabling it to better understand the modified query and generate more accurate answers that align with the user's true intention. Therefore, this stage not only greatly improves the model's performance but also further consolidates the improvements achieved through the previous two fine-tuning stages, making the model more accurate when handling complex tasks.

\subsection{Case Study}
\label{subsec:35}
We provide case studies as examples in Table~\ref{tab:table16} and Appendix~\ref{sec:appendixd}, which visually demonstrate that the model trained with our method can correctly identify and correct the misleading information in the question, compared to the original model, thereby generating the correct answer. This significantly improves the accuracy of the responses and reduces hallucinations.

In the example from the \textit{HaluEval-QA}$_{mis}$ dataset (Table~\ref{tab:table16}), when faced with a query containing misleading information, the original model responds directly based on the misleading question, resulting in an incorrect answer ``FA Premier League''. Although this competition is indeed a British football event, it does not align with the competition that the question actually intends to ask about, which is the one attended by Malcolm Smith. This indicates that misleading information can mislead the model into providing an incorrect response. In contrast, the model trained with our method keenly detects the misleading information in the question, corrects the misleading information, and then generates a response based on the corrected question, yielding the correct result.

\begin{table*}[!ht]
\centering
\small
\begin{tabular}{@{\hspace{5pt}}p{3cm}|p{12cm}@{\hspace{5pt}}}
\toprule
\multicolumn{2}{c}{\textit{HaluEval-QA}$_{mis}$ Example\hspace{0.8cm}Ground Truth Label: Super Bowl XLVIII} \\ \midrule
Query & "knowledge": " Smith was named the Most Valuable Player of Super Bowl XLVIII after they defeated the Denver Broncos. Super Bowl XLVIII was an \textcolor{ourBlue}{American football} game between the American Football Conference (AFC) champion Denver Broncos and National Football Conference (NFC) champion Seattle Seahawks to decide the National Football League (NFL) champion for the 2013 season."

"question": "In which \textcolor{ourRed}{England football} game was Malcolm Smith named Most Valuable player?"
\\ \midrule
Model Output \newline
(Original model) & \textcolor{ourRed}{FA Premier League} \\ \midrule
Model Output \newline
(Our Method) & Malcolm Smith named Most Valuable Player is \textcolor{ourBlue}{American football} game, not \textcolor{ourRed}{England football} game, this American football game is \textcolor{ourBlue}{Super Bowl XLVIII}. \\ \bottomrule
\end{tabular}
\caption{Case study on the \textit{HaluEval-QA}$_{mis}$ dataset. The original model is misled by the misleading information "England football" in the query, resulting in an incorrect answer (highlighted in \textcolor{ourRed}{red}). However, the model trained with our method correctly identifies that it should be ``American football'' and generates the correct answer (highlighted in \textcolor{ourBlue}{blue}).}
\label{tab:table0}
\end{table*}

\subsection{Do the questions in standard datasets contain misleading information?}

In order to investigate whether some potentially misleading information also exists in commonly used standard datasets (original \textit{HaluEval}, \textit{CQA}, and \textit{TruthfulQA}), we employed the Qwen2-7B model, which had been trained in the first stage, to examine the questions from these standard datasets. Our findings revealed that some questions in the standard datasets contain misleading information. The proportions of questions containing misleading information in each dataset are summarized in Table~\ref{tab:misleading}.

Subsequently, we removed the questions identified by the model as containing misleading information from the original datasets to form new datasets (\textit{HaluEval-Sum}$_{sub}$, \textit{HaluEval-Dia}$_{sub}$, \textit{HaluEval-QA}$_{sub}$, \textit{CQA}$_{sub}$). We then conducted evaluations on these new datasets. The experimental results are shown in Table~\ref{tab:comparison}.

We observed that the accuracy of the original Qwen2-7B model's responses on the datasets from which questions containing misleading information had been removed showed a significant improvement compared to the original datasets. This indicates that the misleading information present in the original datasets can interfere with the model's ability to generate accurate responses. However, the Qwen2-7B model trained with our method did not show any notable performance changes, demonstrating its robustness against questions that contain misleading information and its reduced susceptibility to such interference.

\begin{table}[h]
\centering
\begin{tabular}{lc}
\toprule
\textbf{Dataset} & \textbf{Percentage} \\
\midrule
\textit{HaluEval-Sum}         & 3.91\%    \\
\textit{HaluEval-Dia}              & 14.62\%  \\
\textit{HaluEval-QA}               & 13.53\%     \\
\textit{CQA}              & 10.40\%    \\
\textit{TruthfulQA}       & 6.30\%     \\
\bottomrule
\end{tabular}
\caption{The proportion of questions containing misleading information in the three standard datasets.}
\label{tab:misleading}
\end{table}

\begin{table*}[ht]
\setlength\tabcolsep{3.5pt}
\small
\centering
\begin{tabular*}{\textwidth}{@{\extracolsep{\fill}} l *{8}{c} }
\toprule
\multirow{2}{*}{\textbf{Method}} & 
\multicolumn{2}{c}{\textit{HaluEval-Sum}} & 
\multicolumn{2}{c}{\textit{HaluEval-Dia}} & 
\multicolumn{2}{c}{\textit{HaluEval-QA}} & 
\multicolumn{2}{c}{\textit{CQA}} \\
\cmidrule(lr){2-3} \cmidrule(lr){4-5} \cmidrule(lr){6-7} \cmidrule(lr){8-9}
& \textit{Original} & \textit{Sub} & \textit{Original} & \textit{Sub} & \textit{Original} & \textit{Sub} & \textit{Original} & \textit{Sub} \\ 
\midrule
Qwen2-7B
& 51.5±0.8 & 53.6±0.7 
& 70.3±0.4 & 71.1±0.5 
& 54.5±1.3 & 57.2±1.1 
& 70.4±0.8 & 73.1±0.9 \\

\textbf{Ours} 
& \textbf{68.1±1.7} & \textbf{68.5±1.4} 
& \textbf{72.1±1.3} & \textbf{72.3±0.8} 
& \textbf{68.9±1.6} & \textbf{69.7±0.7} 
& \textbf{78.8±1.6} & \textbf{79.1±1.1} \\ 

\bottomrule
\end{tabular*}
\caption{Performance (mean $\pm$ standard deviation, \%) comparison before and after addressing misleading information in the dataset. "Original" denotes the original dataset, and "Sub" refers to the dataset subset with misleading information questions removed.}
\label{tab:comparison}
\end{table*}

\section{Related Work}

\paragraph{The Impact of Misleading Information on LLMs} 
Studies have shown that in question answering systems, when the input contains false or ambiguous information, LLMs often generate inaccurate responses based on these erroneous inputs~\citep{pan2023on,zhang2024clamber,kuhn2022clam,chen2024can}. To address this issue, a common approach is to detect potential errors or inconsistencies based on the syntactic and semantic analysis of the input. For example, \citet{li2024detection} proposed a detection-correction integrated framework based on a general language model. This approach integrates the detection and correction processes into a single model, enabling the automatic detection and correction of syntactic errors in the input text. To ensure the accuracy of the input information, some studies~\citep{DBLP:journals/corr/abs-2410-13788,butala2024promise,xiong2024interactive} have attempted to validate the correctness of the input through multi-turn dialogues, ensuring that the content generated by the model aligns with reality. Although existing methods have improved the model's ability to handle misleading information to some extent, the complexity and diversity of misleading information still pose significant challenges in this area.

\paragraph{Model Output Correction} In recent years, methods aimed at correcting the content generated by LLMs to improve generation accuracy have been widely studied. \citet{wu2024large} proposed an iterative verification-correction framework (ProCo), which identifies and corrects potential errors through subsequent validation of the generated content. \citet{petroni2021kilt} leveraged external knowledge bases to correct the content generated by the model. \citet{liu2024large} explored the intrinsic self-correction capabilities of LLMs from both theoretical and empirical perspectives, proposing that zero temperature and fair prompting are key factors for successful self-correction. \citet{liu2024mind} proposed distilling the LLMs' self-evaluation capability and more comprehensive thinking into smaller models, effectively improving the performance and answer accuracy of LLMs in resource-constrained environments. 
In addition, knowledge-enhanced generation has also been used to reduce the likelihood of language models generating misleading content~\citep{jiang2024disinformation}.

\section{Conclusion}
In this paper, we propose a novel three-stage fine-tuning method to mitigate the impact of misleading information in the input queries on LLMs. Our method enhances the ability of LLMs to detect, correct, and respond accurately to queries containing misleading content, significantly improving response accuracy and reducing hallucinations. Extensive experiments across various datasets demonstrate the effectiveness of this approach, making LLMs more robust and trustworthy.

\section{Limitations}
Although our method has achieved significant improvement in mitigating the impact of misleading information in the input on LLMs, there are still some limitations that can be addressed in future work:
\begin{enumerate}
\item \textbf{Dependency on high-quality knowledge}: Our method relies on the model's internal knowledge or requires reliable external knowledge during the query detection and Query Correction Stages. If both the model's internal knowledge and the external knowledge base are incomplete or inaccurate, the model may still generate incorrect corrections or responses.

\item \textbf{Scalability to larger models}: Although our method performs well across different model sizes (e.g., Qwen-7B vs. Qwen-14B, Llama 3B vs. 8B models), future work can explore experiments on even larger models.

\item \textbf{Expansion of the training dataset}: Future work could consider expanding the training dataset to more effectively improve model performance and facilitate transfer to additional tasks or domains.
\end{enumerate}

\section{Ethical Statement}
\paragraph{The Use of AI Assistants} We employed ChatGPT to assist us in polishing our paper and coding.

\bibliography{custom}

\begin{thebibliography}{37}
\providecommand{\natexlab}[1]{#1}

\bibitem[{Barnard et~al.(2023)Barnard, Van~Sittert, and Rambhatla}]{barnard2023self}
Francois Barnard, Marlize Van~Sittert, and Sirisha Rambhatla. 2023.
\newblock Self-diagnosis and large language models: A new front for medical misinformation.
\newblock \emph{arXiv preprint arXiv:2307.04910}.

\bibitem[{Butala et~al.(2024)Butala, Garg, Banerjee, and Misra}]{butala2024promise}
Yash Butala, Siddhant Garg, Pratyay Banerjee, and Amita Misra. 2024.
\newblock Promise: A proactive multi-turn dialogue dataset for information-seeking intent resolution.
\newblock In \emph{Findings of the Association for Computational Linguistics: EACL 2024}, pages 1774--1789.

\bibitem[{Chen and Shu(2024)}]{chen2024can}
Canyu Chen and Kai Shu. 2024.
\newblock \href {https://openreview.net/forum?id=ccxD4mtkTU} {Can {LLM}-generated misinformation be detected?}
\newblock In \emph{The Twelfth International Conference on Learning Representations}.

\bibitem[{Dubey et~al.(2024)Dubey, Jauhri, Pandey, Kadian, Al-Dahle, Letman, Mathur, Schelten, Yang, Fan et~al.}]{dubey2024llama}
Abhimanyu Dubey, Abhinav Jauhri, Abhinav Pandey, Abhishek Kadian, Ahmad Al-Dahle, Aiesha Letman, Akhil Mathur, Alan Schelten, Amy Yang, Angela Fan, et~al. 2024.
\newblock The llama 3 herd of models.
\newblock \emph{arXiv preprint arXiv:2407.21783}.

\bibitem[{Gao et~al.(2023)Gao, Xiong, Gao, Jia, Pan, Bi, Dai, Sun, and Wang}]{gao2023retrieval}
Yunfan Gao, Yun Xiong, Xinyu Gao, Kangxiang Jia, Jinliu Pan, Yuxi Bi, Yi~Dai, Jiawei Sun, and Haofen Wang. 2023.
\newblock Retrieval-augmented generation for large language models: A survey.
\newblock \emph{arXiv preprint arXiv:2312.10997}.

\bibitem[{Jiang et~al.(2024)Jiang, Tan, Nirmal, and Liu}]{jiang2024disinformation}
Bohan Jiang, Zhen Tan, Ayushi Nirmal, and Huan Liu. 2024.
\newblock Disinformation detection: An evolving challenge in the age of llms.
\newblock In \emph{Proceedings of the 2024 SIAM International Conference on Data Mining (SDM)}, pages 427--435. SIAM.

\bibitem[{Jin et~al.(2024)Jin, Zhang, Jiang, Liu, Liu, Liu, and Jin}]{jin2024ragcache}
Chao Jin, Zili Zhang, Xuanlin Jiang, Fangyue Liu, Xin Liu, Xuanzhe Liu, and Xin Jin. 2024.
\newblock Ragcache: Efficient knowledge caching for retrieval-augmented generation.
\newblock \emph{arXiv preprint arXiv:2404.12457}.

\bibitem[{Kamoi et~al.(2024)Kamoi, Zhang, Zhang, Han, and Zhang}]{kamoi2024can}
Ryo Kamoi, Yusen Zhang, Nan Zhang, Jiawei Han, and Rui Zhang. 2024.
\newblock When can llms actually correct their own mistakes? a critical survey of self-correction of llms.
\newblock \emph{arXiv preprint arXiv:2406.01297}.

\bibitem[{Kuhn et~al.(2022)Kuhn, Gal, and Farquhar}]{kuhn2022clam}
Lorenz Kuhn, Yarin Gal, and Sebastian Farquhar. 2022.
\newblock Clam: Selective clarification for ambiguous questions with generative language models.
\newblock \emph{arXiv preprint arXiv:2212.07769}.

\bibitem[{Lewis et~al.(2020)Lewis, Perez, Piktus, Petroni, Karpukhin, Goyal, K{\"u}ttler, Lewis, Yih, Rockt{\"a}schel et~al.}]{lewis2020retrieval}
Patrick Lewis, Ethan Perez, Aleksandra Piktus, Fabio Petroni, Vladimir Karpukhin, Naman Goyal, Heinrich K{\"u}ttler, Mike Lewis, Wen-tau Yih, Tim Rockt{\"a}schel, et~al. 2020.
\newblock Retrieval-augmented generation for knowledge-intensive nlp tasks.
\newblock \emph{Advances in Neural Information Processing Systems}, 33:9459--9474.

\bibitem[{Li et~al.(2024{\natexlab{a}})Li, Yan, Zhang, Wang, He, Huang, Xue, and Huang}]{li2024role}
Dongyang Li, Junbing Yan, Taolin Zhang, Chengyu Wang, Xiaofeng He, Longtao Huang, Hui Xue, and Jun Huang. 2024{\natexlab{a}}.
\newblock On the role of long-tail knowledge in retrieval augmented large language models.
\newblock In \emph{ACL (Short Papers)}.

\bibitem[{Li et~al.(2023)Li, Cheng, Zhao, Nie, and Wen}]{li2023halueval}
Junyi Li, Xiaoxue Cheng, Wayne~Xin Zhao, Jian-Yun Nie, and Ji-Rong Wen. 2023.
\newblock Halueval: A large-scale hallucination evaluation benchmark for large language models.
\newblock In \emph{Proceedings of the 2023 Conference on Empirical Methods in Natural Language Processing}, pages 6449--6464.

\bibitem[{Li and Wang(2024)}]{li2024detection}
Wei Li and Houfeng Wang. 2024.
\newblock Detection-correction structure via general language model for grammatical error correction.
\newblock \emph{arXiv preprint arXiv:2405.17804}.

\bibitem[{Li et~al.(2024{\natexlab{b}})Li, Xu, Shen, Xu, Gu, Lai, Tao, and Ma}]{li2024leveraging}
Zhen Li, Xiaohan Xu, Tao Shen, Can Xu, Jia-Chen Gu, Yuxuan Lai, Chongyang Tao, and Shuai Ma. 2024{\natexlab{b}}.
\newblock Leveraging large language models for nlg evaluation: Advances and challenges.
\newblock In \emph{Proceedings of the 2024 Conference on Empirical Methods in Natural Language Processing}, pages 16028--16045.

\bibitem[{Lin(2004)}]{lin2004rouge}
Chin-Yew Lin. 2004.
\newblock Rouge: A package for automatic evaluation of summaries.
\newblock In \emph{Text summarization branches out}, pages 74--81.

\bibitem[{Lin et~al.(2022)Lin, Hilton, and Evans}]{lin2022truthfulqa}
Stephanie Lin, Jacob Hilton, and Owain Evans. 2022.
\newblock Truthfulqa: Measuring how models mimic human falsehoods.
\newblock In \emph{Proceedings of the 60th Annual Meeting of the Association for Computational Linguistics (Volume 1: Long Papers)}, pages 3214--3252.

\bibitem[{Liu et~al.(2024{\natexlab{a}})Liu, Nassereldine, Yang, Xu, Hu, Li, Kumar, Lee, and Xiong}]{liu2024large}
Dancheng Liu, Amir Nassereldine, Ziming Yang, Chenhui Xu, Yuting Hu, Jiajie Li, Utkarsh Kumar, Changjae Lee, and Jinjun Xiong. 2024{\natexlab{a}}.
\newblock Large language models have intrinsic self-correction ability.
\newblock \emph{CoRR}.

\bibitem[{Liu et~al.(2024{\natexlab{b}})Liu, Li, Zhang, Du, Chen, Hu, Xu, Chen, and Wu}]{liu2024mind}
Weize Liu, Guocong Li, Kai Zhang, Bang Du, Qiyuan Chen, Xuming Hu, Hongxia Xu, Jintai Chen, and Jian Wu. 2024{\natexlab{b}}.
\newblock Mind’s mirror: Distilling self-evaluation capability and comprehensive thinking from large language models.
\newblock In \emph{Proceedings of the 2024 Conference of the North American Chapter of the Association for Computational Linguistics: Human Language Technologies (Volume 1: Long Papers)}, pages 6748--6763.

\bibitem[{Madaan et~al.(2024)Madaan, Tandon, Gupta, Hallinan, Gao, Wiegreffe, Alon, Dziri, Prabhumoye, Yang et~al.}]{madaan2024self}
Aman Madaan, Niket Tandon, Prakhar Gupta, Skyler Hallinan, Luyu Gao, Sarah Wiegreffe, Uri Alon, Nouha Dziri, Shrimai Prabhumoye, Yiming Yang, et~al. 2024.
\newblock Self-refine: Iterative refinement with self-feedback.
\newblock \emph{Advances in Neural Information Processing Systems}, 36.

\bibitem[{Nie et~al.(2020)Nie, Williams, Dinan, Bansal, Weston, and Kiela}]{nie2020adversarial}
Yixin Nie, Adina Williams, Emily Dinan, Mohit Bansal, Jason Weston, and Douwe Kiela. 2020.
\newblock Adversarial nli: A new benchmark for natural language understanding.
\newblock In \emph{Proceedings of the 58th Annual Meeting of the Association for Computational Linguistics}, pages 4885--4901.

\bibitem[{Niu et~al.(2018)Niu, Qiao, Li, and Huang}]{niu2018word}
Yilin Niu, Chao Qiao, Hang Li, and Minlie Huang. 2018.
\newblock Word embedding based edit distance.
\newblock \emph{arXiv preprint arXiv:1810.10752}.

\bibitem[{Pan et~al.(2023)Pan, Pan, Chen, Nakov, Kan, and Wang}]{pan2023on}
Yikang Pan, Liangming Pan, Wenhu Chen, Preslav Nakov, Min-Yen Kan, and William~Yang Wang. 2023.
\newblock \href {https://openreview.net/forum?id=voBhcwDyPt} {On the risk of misinformation pollution with large language models}.
\newblock In \emph{The 2023 Conference on Empirical Methods in Natural Language Processing}.

\bibitem[{Papineni et~al.(2002)Papineni, Roukos, Ward, and Zhu}]{papineni2002bleu}
Kishore Papineni, Salim Roukos, Todd Ward, and Wei-Jing Zhu. 2002.
\newblock Bleu: a method for automatic evaluation of machine translation.
\newblock In \emph{Proceedings of the 40th annual meeting of the Association for Computational Linguistics}, pages 311--318.

\bibitem[{Petroni et~al.(2021)Petroni, Piktus, Fan, Lewis, Yazdani, Cao, Thorne, Jernite, Karpukhin, Maillard et~al.}]{petroni2021kilt}
F~Petroni, A~Piktus, A~Fan, PSH Lewis, M~Yazdani, ND~Cao, J~Thorne, Y~Jernite, V~Karpukhin, J~Maillard, et~al. 2021.
\newblock Kilt: a benchmark for knowledge intensive language tasks.
\newblock In \emph{NAACL-HLT}, pages 2523--2544. Association for Computational Linguistics.

\bibitem[{Reimers(2019)}]{reimers2019sentence}
N~Reimers. 2019.
\newblock Sentence-bert: Sentence embeddings using siamese bert-networks.
\newblock \emph{arXiv preprint arXiv:1908.10084}.

\bibitem[{Sellam et~al.(2020)Sellam, Das, and Parikh}]{sellam2020bleurt}
Thibault Sellam, Dipanjan Das, and Ankur~P Parikh. 2020.
\newblock Bleurt: Learning robust metrics for text generation.
\newblock \emph{arXiv preprint arXiv:2004.04696}.

\bibitem[{Song et~al.(2024)Song, Yu, and Yoon}]{song2024large}
Jongyoon Song, Sangwon Yu, and Sungroh Yoon. 2024.
\newblock Large language models are skeptics: False negative problem of input-conflicting hallucination.
\newblock \emph{arXiv preprint arXiv:2406.13929}.

\bibitem[{Talmor et~al.(2019)Talmor, Herzig, Lourie, and Berant}]{talmor2019commonsenseqa}
Alon Talmor, Jonathan Herzig, Nicholas Lourie, and Jonathan Berant. 2019.
\newblock Commonsenseqa: A question answering challenge targeting commonsense knowledge.
\newblock In \emph{Proceedings of the 2019 Conference of the North American Chapter of the Association for Computational Linguistics: Human Language Technologies, Volume 1 (Long and Short Papers)}, pages 4149--4158.

\bibitem[{Wu et~al.(2022)Wu, Xin, Chen, Han, and Sun}]{wu2022semantic}
Shan Wu, Chunlei Xin, Bo~Chen, Xianpei Han, and Le~Sun. 2022.
\newblock Semantic-aware contrastive learning for more accurate semantic parsing.
\newblock In \emph{Proceedings of the 2022 Conference on Empirical Methods in Natural Language Processing}, pages 4040--4052.

\bibitem[{Wu et~al.(2024{\natexlab{a}})Wu, Xie, Chen, Zhu, Zhang, and Xiao}]{wu2024easily}
Siye Wu, Jian Xie, Jiangjie Chen, Tinghui Zhu, Kai Zhang, and Yanghua Xiao. 2024{\natexlab{a}}.
\newblock How easily do irrelevant inputs skew the responses of large language models?
\newblock \emph{arXiv preprint arXiv:2404.03302}.

\bibitem[{Wu et~al.(2024{\natexlab{b}})Wu, Zeng, Zhang, Tan, Shen, and Jiang}]{wu2024large}
Zhenyu Wu, Qingkai Zeng, Zhihan Zhang, Zhaoxuan Tan, Chao Shen, and Meng Jiang. 2024{\natexlab{b}}.
\newblock Large language models can self-correct with key condition verification.
\newblock In \emph{Proceedings of the 2024 Conference on Empirical Methods in Natural Language Processing}, pages 12846--12867.

\bibitem[{Xiong et~al.(2024)Xiong, Bao, and Zhao}]{xiong2024interactive}
Guanming Xiong, Junwei Bao, and Wen Zhao. 2024.
\newblock Interactive-kbqa: Multi-turn interactions for knowledge base question answering with large language models.
\newblock \emph{arXiv preprint arXiv:2402.15131}.

\bibitem[{Xuanfan and Piji(2023)}]{xuanfan2023systematic}
Ni~Xuanfan and Li~Piji. 2023.
\newblock A systematic evaluation of large language models for natural language generation tasks.
\newblock In \emph{Proceedings of the 22nd Chinese National Conference on Computational Linguistics (Volume 2: Frontier Forum)}, pages 40--56.

\bibitem[{Yang et~al.(2024)Yang, Yang, Zhang, Hui, Zheng, Yu, Li, Liu, Huang, Wei et~al.}]{yang2024qwen2}
An~Yang, Baosong Yang, Beichen Zhang, Binyuan Hui, Bo~Zheng, Bowen Yu, Chengyuan Li, Dayiheng Liu, Fei Huang, Haoran Wei, et~al. 2024.
\newblock Qwen2. 5 technical report.
\newblock \emph{arXiv preprint arXiv:2412.15115}.

\bibitem[{Yu et~al.(2024)Yu, Zhang, and Feng}]{yu2024truth}
Tian Yu, Shaolei Zhang, and Yang Feng. 2024.
\newblock Truth-aware context selection: Mitigating the hallucinations of large language models being misled by untruthful contexts.
\newblock \emph{arXiv preprint arXiv:2403.07556}.

\bibitem[{Zhang et~al.(2024{\natexlab{a}})Zhang, Knox, and Choi}]{DBLP:journals/corr/abs-2410-13788}
Michael J.~Q. Zhang, W.~Bradley Knox, and Eunsol Choi. 2024{\natexlab{a}}.
\newblock \href {https://doi.org/10.48550/arXiv.2410.13788} {Modeling future conversation turns to teach llms to ask clarifying questions}.
\newblock \emph{CoRR}, abs/2410.13788.

\bibitem[{Zhang et~al.(2024{\natexlab{b}})Zhang, Qin, Deng, Huang, Lei, Liu, Jin, Liang, and Chua}]{zhang2024clamber}
Tong Zhang, Peixin Qin, Yang Deng, Chen Huang, Wenqiang Lei, Junhong Liu, Dingnan Jin, Hongru Liang, and Tat-Seng Chua. 2024{\natexlab{b}}.
\newblock Clamber: A benchmark of identifying and clarifying ambiguous information needs in large language models.
\newblock \emph{arXiv e-prints}, pages arXiv--2405.

\end{thebibliography}

\clearpage
\appendix

\section{Experimental details}
\label{sec:appendixa}

We employ LoRA for efficient fine-tuning. The detailed setting of hyperparameters is shown in Table~\ref{tab:table6}. Table~\ref{tab:table7} presents the information about the datasets used for training and testing. Regarding SFT, we use the \textit{QA} and \textit{HaluEval-QA}$_{mis}$ datasets for implementation.
\begin{table}[h]
\centering
\begin{tabular}{lc}
\toprule
\textbf{Configuration} & \textbf{Value} \\
\midrule
Number of epochs         & 4    \\
Devices             & 4 Nvidia H20 GPU (80G)  \\
Total Batch size    & 128     \\
Learning rate       & $5 \times 10^{-4}$    \\
Warmup Ratio           & 0.1     \\
\bottomrule
\end{tabular}
\caption{Finetuning hyperparameters for experiments.}
\label{tab:table6}
\end{table}

\begin{table}[ht]
\small
\centering
\begin{tabular}{lcc}
\toprule
\textbf{Datasets} & \textbf{Train number} & \textbf{Test number} \\
\midrule
\textit{HaluEval-QA}            & 4000          & 6000\\
\textit{HaluEval-QA}$_{mis}$      & 4000          & 6000\\
\textit{HaluEval-Sum} & 0          & 10000\\
\textit{HaluEval-Dia}      & 0          & 10000\\
\textit{CQA}           & 0          & 1221\\
\textit{CQA}$_{mis}$     & 0          & 1221\\
\textit{TruthfulQA}    & 0          & 818\\
\bottomrule
\end{tabular}
\caption{Dataset Split.}
\label{tab:table7}
\end{table}

\section{Dataset details}
\label{sec:appendixb}

The specific process of constructing our training dataset is as follows:
\begin{itemize}
    \item \textbf{Query Detection Stage:}  
    We used a mixed dataset comprising \textit{QA} and \textit{HaluEval-QA}$_{mis}$ samples for training. Additionally, we introduced a field $with-misleading$ as a label to indicate whether a query contains misleading information. For queries in the \textit{QA} dataset, the $with\text{-}misleading$ field is set to ``NO'', indicating that the query does not contain misleading information. For queries in the \textit{HaluEval-QA}$_{mis}$ dataset, the $with\text{-}misleading$ field is set to ``YES'', indicating that the query contains misleading information.  

    \item \textbf{Query Correction Stage:}  
    Since the task in this stage requires correcting queries that contain misleading information, we selected 4,000 samples from the \textit{HaluEval-QA}$_{mis}$ dataset for training. At this point, our training objective is to correct the misleading queries and transform them into the corresponding queries from the original \textit{QA} dataset. 

    \item \textbf{Answer Generation Stage:} In this stage, we also use a mixed dataset of \textit{QA} and \textit{HaluEval-QA}$_{mis}$ samples to train the model to generate correct answers, where the queries containing misleading information have already been corrected in the query correction stage.
\end{itemize}

This dataset construction strategy ensures that the model can focus on the corresponding data characteristics at each task stage, while also enhancing the model's ability to handle diverse queries. Next, we present the prompts that we used in the experiments as well as the \textit{HaluEval-QA}$_{mis}$ and \textit{CQA}$_{mis}$ datasets. Table~\ref{tab:table8} and Table~\ref{tab:table10} show the templates used to generate the \textit{HaluEval-QA}$_{mis}$ and \textit{CQA}$_{mis}$ datasets, while Table~\ref{tab:table9} and Table~\ref{tab:table11} display some data from the \textit{HaluEval-QA}$_{mis}$ and \textit{CQA}$_{mis}$ datasets. Table~\ref{tab:table12}, Table~\ref{tab:table13}, and Table~\ref{tab:table14} respectively show the templates for query detection, misleading query correction, and answer generation.

\begin{table*}[ht]
\small
\raggedright
\begin{tabularx}{\textwidth}{X}
\toprule
\textbf{Prompt for Generating \textit{HaluEval-QA}$_{mis}$ Dataset} \\
\midrule
You are a language expert. You should create a question that includes subtle mistakes derived from the content in the knowledge.

The modified question should retain the general meaning of the original but introduce small errors or misdirecting details such as incorrect entity names, relationships, times, or terminology. These changes should be subtle and should not drastically alter the original question's structure or core intent.

Do not introduce completely new concepts, terms, or laws that may cause the mistake question to deviate too far from the original. The mistake question should still be recognizable as a version of the original question, with enough misleading elements to potentially cause an incorrect answer if not identified and corrected.

The goal is to create a subtle version of the original question that misleads the model, but does not make it so different that the answer is hard to recover.

Do not provide any content, only the misleading question.
\vspace{1\baselineskip}

"""

\#knowledge\#: \{knowledge\}

\#question\#: \{question\}

""" \\
\bottomrule
\end{tabularx}
\caption{The prompt for generating the \textit{HaluEval-QA}$_{mis}$ dataset.}
\label{tab:table8}
\end{table*}

\begin{table*}[ht]
\small
\raggedright %
\begin{tabularx}{\textwidth}{X} %
\toprule
\textbf{The example of \textit{HaluEval-QA}$_{mis}$ dataset} \\
\midrule

\textbf{Example1:}

"knowledge": "The 2003 SEC Championship Game was won by the LSU Tigers 34–13 over the Georgia Bulldogs. The game was played in the \textcolor{ourBlue}{Georgia Dome in Atlanta, Georgia}. Located in Atlanta, Georgia, between downtown to the east and Vine City to the west, it was owned and operated by the State of Georgia as part of the Georgia World Congress Center Authority."

"question": "The 2003 SEC Championship Game was played at a stadium owned and operated by who?"

"misleading question": "The 2003 SEC Championship Game was played at a stadium \textcolor{ourRed}{in the United States} owned and operated by who?"

"right answer": "the State of Georgia as part of the Georgia World Congress Center Authority"

"hallucinated answer": "The film \"Pandorum\" was released in 2010." 
"The 2003 SEC Championship Game was played at the Mercedes-Benz Stadium in Atlanta."\\
\midrule

\textbf{Example2:}

"knowledge": " He was also a successful author, writing biographies of politician Al Smith and athletes Lou Gehrig and John McGraw, as well as histories of the New York Yankees, New York Giants and Brooklyn Dodgers. The team's name derived from the reputed skill of \textcolor{ourBlue}{Brooklyn residents} at evading the city's trolley streetcar network."

"question": "What  team's name derived from the reputed skill of \textcolor{ourBlue}{Brooklyn residents} at evading the city's trolley streetcar network, who had a history wrote of it by an American sportswriter and biographer who overed sports in New York for the \"New York Sun\" from 1915 to 1943?"

"misleading question": "What team's name derived from the reputed skill of \textcolor{ourRed}{Manhattan residents} at evading the city's trolley streetcar network?"

"right answer": "Brooklyn Dodgers"

"hallucinated answer": "The New York Yankees were named after the reputed skill of Brooklyn residents at evading the city's trolley streetcar network." \\
\midrule

\textbf{Example3:}

"knowledge": "Sooni Taraporevala (born 1957) is an Indian screenwriter, photographer and filmmaker who is best known as the screenwriter of \"Mississippi Masala\", \"The Namesake\" and Oscar-nominated \"Salaam Bombay\" (1988), all directed by Mira Nair. Mississippi Masala is a \textcolor{ourBlue}{1991 romantic drama} film directed by Mira Nair, based upon a screenplay by Sooni Taraporevala, starring Denzel Washington, Sarita Choudhury, and Roshan Seth."

"question": "Which Oscar-nominated film was written by the screenwriter of a \textcolor{ourRed}{1990 romantic drama} starring Sarita Choudhury?"

"right answer": "Salaam Bombay"

"hallucinated answer": "The Namesake" \\
\bottomrule
\end{tabularx}
\caption{Randomly sampled \textit{HaluEval-QA}$_{mis}$ dataset examples. \textcolor{ourBlue}{Blue} represents correct information, while \textcolor{ourRed}{red} represents misleading information.}
\label{tab:table9}
\end{table*}

\begin{table*}[ht]
\raggedright %
\small
\begin{tabularx}{\textwidth}{X} %
\toprule
\textbf{Prompt for generating \textit{CQA}$_{mis}$ dataset} \\
\midrule
You are a language expert. Your task is to take a given question and modify it by introducing distracting or irrelevant details to make the problem more complex and harder to answer. These distractions should not directly lead to any of the answer choices, especially the correct one.

Steps:

1. Modify the Question: Add irrelevant or distracting details that make the question more challenging. These distractions can be excessive descriptions, background information, or other elements that do not directly relate to the core of the question. 

2. Provide a Detailed Explanation: After modifying the question, provide a clear explanation that shows why the correct answer is still valid. The explanation should demonstrate that the added distractions don’t affect the reasoning behind the correct answer and why it is still the best choice.
 
Important Notes:
The distractions should increase the complexity of the question, but should not directly or indirectly guide the reader to the correct answer.
The explanation should clearly outline why the correct answer is the best choice and how the added distractions do not change the fundamental reasoning.
\vspace{1\baselineskip}

Example:

"""

"question": "A thoroughfare meandered through fields and woods, where was it passing though?",

"choices": "["move about","city","country","town","new york city"]".

"""
\vspace{1\baselineskip}

"mistake question": "This road winds through various landscapes, sometimes open farmlands, other times dense forests, and occasionally small villages. Along the way, there are signs of modern construction like small factories and even a few billboards. The surroundings shift frequently, with the occasional sight of tall buildings in the distance. Where is this road most likely passing through?"

"explanation": "The road is passing through country because, despite the modern construction and occasional tall buildings in the distance, the key elements—farmlands, forests, and small villages are characteristic of rural, countryside areas. The mention of "small factories" and "billboards" are distractions that suggest a more urban environment, but they do not outweigh the overall rural characteristics of the area described."
\vspace{1\baselineskip}

"""

\#question\#: \{question\}

\#choices\#: \{choices\}

""" \\
\bottomrule
\end{tabularx}
\caption{The prompt for generating the \textit{CQA}$_{mis}$ dataset. In addition to the original fields, we also generated an explain field to provide a detailed description of how the question, after adding misleading information, is step by step deduced to the correct answer.}
\label{tab:table10}
\end{table*}

\begin{table*}[ht]
\raggedright %
\small
\begin{tabularx}{\textwidth}{X} %
\toprule
\textbf{The example of \textit{CQA}$_{mis}$ dataset} \\
\midrule

\textbf{Example1:}

"question": "Billy was reading the newspaper as he commuted to work, but once he got to his destination he balled it up and put it somewhere. Where did it put it?"

"choices": ["trash", "floor", "subway", "ground", "lawn"]

"answer": "trash"

"misleading question": Billy was engrossed in reading the morning newspaper during his daily commute. As he arrived at his workplace, \textcolor{ourRed}{he noticed the crowded train and decided to quickly stuff the newspaper into his pocket to avoid carrying it around.} However, as the day progressed, he realized he had forgotten about it and didn't know where he had placed it. Where could it be now?"

"explanation": Despite the added details of Billy stuffing the newspaper into his pocket and forgetting about it later, the core question remains about where Billy put the newspaper once he reached his destination. The hallucinations do not provide any information about the newspaper's final location; thus, the correct answer, \"trash,\" remains valid. Billy might have thrown away the newspaper if he didn't need it anymore or if he found it cluttering his belongings." \\
\midrule
\textbf{Example2:}

"question": "He had to wear a tuxedo while playing the keyboard instrument, so did the other hundred members of the what?"

"choices": ["music store", "band", "medium", "orchestra", "piano store"]

"answer": "orchestra"

"misleading question": "He had to wear a tuxedo while playing the keyboard instrument, and the other hundred members also wore tuxedos. \textcolor{ourRed}{There was a piano beside him.} Which group did they belong to?"

"explanation": "Although there was a piano beside him, the fact that they had to wear tuxedos still leads to the correct answer being "orchestra." \\
\bottomrule
\end{tabularx}
\caption{Randomly sampled \textit{CQA}$_{mis}$ dataset examples. \textcolor{ourRed}{Red} represents misleading information.}
\label{tab:table11}
\end{table*}

\begin{table*}[ht]
\raggedright %
\small
\begin{tabularx}{\textwidth}{X} %
\toprule
\textbf{Prompt for question detection} \\
\midrule

You are an excellent mistake detector.

I will provide you with a \#knowledge\# and a \#question\#. Your task is to carefully compare the \#knowledge\# and the \#question\#, and based on the content of the \#knowledge\#, determine whether the \#question\# contains any mistake.

If it contains a mistake, output 'YES', otherwise output 'NO'. Please make sure there are no extra outputs.

The following is an example:
\vspace{1\baselineskip}

\#knowledge\#: "The area covered by South East Queensland varies, depending on the definition of the region, though it tends to include Queensland's three largest cities: the capital city Brisbane; the Gold Coast; and the Sunshine Coast. The Gold Coast is a coastal area in the Australian state of Queensland, approximately 66 km south-southeast of the state capital Brisbane and immediately north of the border with New South Wales.".

\#question\#: "South East Queensland (SEQ) is a bio-geographical, political, and administrative region of the state of Queensland in Australia, the area covered by South East Queensland varies, depending on the definition of the region, though it tends to include Queensland's three largest cities, including which coastal area in the Australian state of Victoria, approximately 66 km south-southeast of the state capital Melbourne and immediately north of the border with New South Wales?".

\#your choice\#:"YES"
\vspace{1\baselineskip}

You need to do your best to identify whether there is any content in the \#question\# that contradicts the \#knowledge\#, and if there is, output "YES", otherwise output "NO". \\
\bottomrule
\end{tabularx}
\caption{The prompt for question detection.}
\label{tab:table12}
\end{table*}

\begin{table*}[ht]
\raggedright %
\small
\begin{tabularx}{\textwidth}{X} %
\toprule
\textbf{Prompt for misleading question correction} \\
\midrule

You are an excellent mistake corrector.
I will provide you with a \#knowledge\# and a \#question\#. Your task is to carefully compare the \#knowledge\# and the \#question\#, and based on the content of the \#knowledge\# and real-world information, correct any mistakes in the \#question\# and output the corrected content.

The following is an example:
\vspace{1\baselineskip}

\#knowledge\#: "The area covered by South East Queensland varies, depending on the definition of the region, though it tends to include Queensland's three largest cities: the capital city Brisbane; the Gold Coast; and the Sunshine Coast. The Gold Coast is a coastal area in the Australian state of Queensland, approximately 66 km south-southeast of the state capital Brisbane and immediately north of the border with New South Wales."

\#question\#: "South East Queensland (SEQ) is a bio-geographical, political, and administrative region of the state of Queensland in Australia, the area covered by South East Queensland varies, depending on the definition of the region, though it tends to include Queensland's three largest cities, including which coastal area in the Australian state of Victoria, approximately 66 km south-southeast of the state capital Melbourne and immediately north of the border with New South Wales?".

\#your answer\#: "South East Queensland (SEQ) is a bio-geographical, political, and administrative region of the state of Queensland in Australia, the area covered by South East Queensland varies, depending on the definition of the region, though it tends to include Queensland's three largest cities, including which coastal area in the Australian state of Queensland, approximately 66 km south-southeast of the state capital Brisbane and immediately north of the border with New South Wales?"
\vspace{1\baselineskip}

You need to carefully identify if there is any content in the \#question\# that contradicts the \#knowledge\# or real-world knowledge. If there is, please modify the \#question\# to the correct form and output it. Otherwise, directly output the original \#question\#. \\
\bottomrule
\end{tabularx}
\caption{The prompt for misleading question correction.}
\label{tab:table13}
\end{table*}

\begin{table*}[ht]
\raggedright %
\small
\begin{tabularx}{\textwidth}{X} %
\toprule
\textbf{Prompt for generate answer} \\
\midrule

You are an expert with broad knowledge.
I will provide \#knowledge\# and \#question\#. Your task is to answer the \#question\# based on the \#knowledge\# and real-world information, and output the answer to the \#question\#.

The following is an example:
\vspace{1\baselineskip}

\#knowledge\#: "He was one of the pioneers of gambling in Las Vegas, where he was a partner with flamboyant mobster Bugsy Siegel at the Flamingo Hotel. Benjamin \"Bugsy\" Siegel (February 28, 1906 – June 20, 1947) was a Jewish American mobster. Siegel was not only influential within the Jewish mob but, like his friend and fellow gangster Meyer Lansky, he also held significant influence within the Italian-American Mafia and the largely Italian-Jewish National Crime Syndicate.".

\#question\#: "What hotel did the influential National Crime Syndicate member found with David Berman?".

\#your answer\#:"Flamingo Hotel"
\vspace{1\baselineskip}

You need to ensure that the answer you provide contains no content that contradicts the \#knowledge\# or real-world knowledge." \\
\bottomrule
\end{tabularx}
\caption{The prompt for generating an answer.}
\label{tab:table14}
\end{table*}

\section{Select misleading question}
\label{sec:appendixc}
This section demonstrates how to select high-quality misleading questions using sentence similarity and answer error rate, with the detailed process shown in Table~\ref{tab:table15}.

\begin{table*}[!ht]
\centering
\small
\begin{tabular}{@{\hspace{5pt}}p{3cm}|p{12cm}@{\hspace{5pt}}}
\toprule
\multicolumn{2}{c}{\textit{HaluEval-QA}$_{mis}$ Example\hspace{0.8cm}Ground Truth Label:1965} \\ \midrule
\textbf{Knowledge} & DuPont v. Kolon Industries is an intellectual property lawsuit centering on the allegation that Kolon Industries (of ), a South Korea-based company, stole trade secrets concerning the production and marketing of Kevlar from DuPont, an American chemical company. Kevlar is a high strength synthetic fiber used in applications as diverse as bicycle tires and body armor. Developed by Stephanie Kwolek at DuPont in 1965, this high-strength material was first commercially used in the early 1970s as a replacement for steel in racing tires.
\\ \midrule
\textbf{Original Query} & What year was the high strength synthetic fiber, which was the subject of a lawsuit between DuPont and Kolon Industries, developed? \\ \midrule
\textbf{Misleading Query 1} & What year was the \textcolor{ourRed}{low strength} synthetic fiber, which was the subject of a lawsuit between DuPont and Kolon Industries, developed? \newline \textbf{Similarity Score:0.96} \newline respond1: 1867 (\textcolor{ourRed}{\ding{56}}) \newline respond2: 1965 (\textcolor{ourBlue}{\ding{52}}) \newline respond3: 1433 (\textcolor{ourRed}{\ding{56}}) \newline \textbf{Error Rate:0.67} \\ \midrule
\textbf{Misleading Query 2} & What year was the high strength synthetic fiber, which was the subject of a lawsuit between DuPont and \textcolor{ourRed}{Japan} Industries, developed? \newline \textbf{Similarity Score:0.96} \newline respond1: 1965 (\textcolor{ourBlue}{\ding{52}}) \newline respond2: 1965 (\textcolor{ourBlue}{\ding{52}}) \newline respond3: 1965 (\textcolor{ourBlue}{\ding{52}}) \newline \textbf{Error Rate:0} 
\\ \midrule
\textbf{Misleading Query 3} & What year was the high strength synthetic fiber, which was the subject of a lawsuit between \textcolor{ourRed}{China and England} Industries, developed? \newline \textbf{Similarity Score:0.92} \newline respond1: 1965 (\textcolor{ourBlue}{\ding{52}}) \newline respond2: 1965 (\textcolor{ourBlue}{\ding{52}}) \newline respond3: 1563 (\textcolor{ourRed}{\ding{56}}) \newline \textbf{Error Rate:0.33} 
\\ \bottomrule
\end{tabular}
\caption{An example of randomly generated misleading queries and the calculation of $S_{\text{sim}}$ and $E_{\text{error}}$, where the misleading information is highlighted in \textcolor{ourRed}{red}. Example 1 satisfies our filtering condition of $S_{\text{sim}}$ > 0.8 and $E_{\text{error}}$ > 0.5. Correct answers are marked with a correct mark \textcolor{ourBlue}{\ding{52}}, while incorrect answers are marked with a wrong mark \textcolor{ourRed}{\ding{56}}. Among them, misleading query 1 satisfies our filtering criteria of $S_{\text{sim}}$ > 0.8 and $E_{\text{error}}$ > 0.5.}
\label{tab:table15}
\end{table*}

\section{Case study}
\label{sec:appendixd}

Similar to the analysis in Section~\ref{subsec:35}, in the example of the \textit{CQA}$_{mis}$ dataset (Table~\ref{tab:table17}), the query mentions that after the mechanic made some adjustments to the engine, the car started to experience slight vibrations and emit some unusual sounds. These distracting pieces of information can easily lead the model to mistakenly interpret the sounds as coming from the horn, resulting in an incorrect answer ``honk the horn''. However, the model trained with our method effectively filters out the interference from the misleading information and provides a response stating that the `unusual sounds' and `slight vibrations' are simply distractions that might suggest the car is malfunctioning, but the fact that it’s operational and no serious issues are mentioned indicates the car will start running as expected after repairs.

We present a case study of the query correction stage in Table~\ref{tab:table5}, the model trained using our method is able to quickly detect misleading information in the question. By leveraging the knowledge from ``the last man executed in Australia'', it correctly modifies the misleading information in the query, ``the last person executed in New Zealand'', and generates the correct question.

\begin{table*}[!ht]
\centering
\small
\begin{tabular}{@{\hspace{5pt}}p{3cm}|p{12cm}@{\hspace{5pt}}}
\toprule
\multicolumn{2}{c}{\textit{HaluEval-QA}$_{mis}$ Example\hspace{0.8cm}Ground Truth Label: Super Bowl XLVIII} \\ \midrule
Query & "knowledge": " Smith was named the Most Valuable Player of Super Bowl XLVIII after they defeated the Denver Broncos. Super Bowl XLVIII was an \textcolor{ourBlue}{American football} game between the American Football Conference (AFC) champion Denver Broncos and National Football Conference (NFC) champion Seattle Seahawks to decide the National Football League (NFL) champion for the 2013 season."

"question": "In which \textcolor{ourRed}{England football} game was Malcolm Smith named Most Valuable player?"
\\ \midrule
Model Output \newline
(Original model) & \textcolor{ourRed}{FA Premier League} \\ \midrule
Model Output \newline
(Our Method) & Malcolm Smith named Most Valuable Player is \textcolor{ourBlue}{American football} game, not \textcolor{ourRed}{England football} game, this American football game is \textcolor{ourBlue}{Super Bowl XLVIII}. \\ \bottomrule
\end{tabular}
\caption{A case study on the \textit{HaluEval-QA}$_{mis}$ dataset. The original model is misled by the misleading information "England football" in the question, resulting in an incorrect answer (highlighted in \textcolor{ourRed}{red}). However, the model trained with our method correctly identifies that it should be ``American football'' and generates the correct answer (highlighted in \textcolor{ourBlue}{blue}).}
\label{tab:table16}
\end{table*}

\begin{table*}[!ht]
\centering
\small
\begin{tabular}{@{\hspace{5pt}}p{3cm}|p{12cm}@{\hspace{5pt}}}
\toprule
\multicolumn{2}{c}{\textit{CQA}$_{mis}$ Example\hspace{0.8cm}Ground Truth Label: start running} \\ \midrule
Query & "question": "After the mechanic performed several adjustments to the engine, including checking the fuel system and replacing some critical parts, the car started to behave differently. It began to \textcolor{ourRed}{vibrate slightly and made some unusual sounds}. Although it seems operational, it's not the same as it was before. What is it most likely to do next?"

"choices": "["going too fast","last several years","honk the horn","go fast","start running"]"
\\ \midrule
Model Output \newline
(Original model) & \textcolor{ourRed}{honk the horn} \\ \midrule
Model Output \newline
(Our Method) & Even though the car is vibrating and making unusual sounds, these are typical signs of a car being started after maintenance, especially if adjustments were made to the engine. The "unusual sounds" and "slight vibrations" are simply distractions that might suggest the car is malfunctioning, but the fact that \textcolor{ourBlue}{it’s operational and no serious issues are mentioned} indicates the car will \textcolor{ourBlue}{start running} as expected after repairs. \\ \bottomrule
\end{tabular}
\caption{A case study on the \textit{CQA}$_{mis}$ dataset. The original model is misled by the misleading information "slight vibrations" and "unusual sounds" in the question, resulting in an incorrect answer (highlighted in \textcolor{ourRed}{red}). However, the model trained with our method correctly identifies "it’s operational and no serious issues are mentioned" and generates the correct answer "the car will start running" (highlighted in \textcolor{ourBlue}{blue}).}
\label{tab:table17}
\end{table*}

\begin{table*}[!ht]
\centering
\small
\begin{tabular}{@{\hspace{5pt}}p{3cm}|p{12cm}@{\hspace{5pt}}}
\toprule
\textbf{Category} & \textbf{Content} \\
\midrule
Knowledge Input & His most well-known work is the award-winning stage play 'Remember Ronald Ryan', a dramatization of the life and subsequent death of Ronald Ryan, the last man executed in Australia. Ronald Joseph Ryan (21 February 1925 – 3 February 1967) was the last person to be legally executed in Australia. \\
\midrule
Original Question & Barry Dickins wrote a film about the last person executed in \textcolor{ourRed}{New Zealand}, named what? \\
\midrule
Output & The knowledge mentions the last person legally executed in \textcolor{ourBlue}{Australia}, not \textcolor{ourRed}{New Zealand}, so the question should be modified to "Barry Dickins wrote a film about the last person executed in \textcolor{ourBlue}{Australia}, named what?" \\
\bottomrule
\end{tabular}
\caption{An example of correcting misleading inputs based on knowledge. For the input data, we used the same prompt as in the query correction training stage (Table~\ref{tab:table13}). The information marked in \textcolor{ourBlue}{blue} represents correct information, while the information marked in \textcolor{ourRed}{red} represents misleading information.}
\label{tab:table5}
\end{table*}

\end{document}